\documentclass[10pt,twocolumn,letterpaper]{article}

\usepackage{iccv}
\usepackage{times}
\usepackage{epsfig}
\usepackage{graphicx}
\usepackage{amsmath}
\usepackage{amssymb}

\usepackage{soul}
\usepackage{diagbox}
\usepackage{multirow}

\usepackage[breaklinks=true,bookmarks=false]{hyperref}

\iccvfinalcopy 

\ificcvfinal\fi

\begin{document}
\title{PU-EVA: An Edge Vector based Approximation Solution for Flexible-scale\\ Point Cloud Upsampling}

\author{
	Luqing Luo$^{1}$, Lulu Tang$^{1}$\thanks{Equal contributions as co-first author}, Wanyi Zhou$^2$, Shizheng Wang$^3$, Zhi-Xin Yang$^1$\thanks{Corresponding author}\\
	$^1$State Key Laboratory of Internet of Things for Smart City, University of Macau\\
	$^2$South China University of Technology\\
	$^3$Institute of Microelectronics Chinese Academy of Sciences\\
	{\tt\small \{gabrielle\_tse, lulutamg\}@outlook.com, 202021044254@mail.scut.edu.cn,} \\
	{\tt\small shizheng.wang@foxmail.com, zxyang@um.edu.mo}
}
\maketitle

\ificcvfinal\thispagestyle{empty}\fi

\begin{abstract}
High-quality point clouds have practical significance for point-based rendering, semantic understanding, and surface reconstruction. Upsampling sparse, noisy and nonuniform point clouds for a denser and more regular approximation of target objects is a desirable but challenging task. Most existing methods duplicate point features for upsampling, constraining the upsampling scales at a fixed rate. In this work, the flexible upsampling rates are achieved via edge vector based affine combinations, and a novel design of Edge Vector based Approximation for Flexible-scale Point clouds Upsampling (PU-EVA) is proposed. The edge vector based approximation encodes the neighboring connectivity via affine combinations based on edge vectors, and restricts the approximation error within the second-order term of Taylor’s Expansion. The EVA upsampling decouples the upsampling scales with network architecture, achieving the flexible upsampling rates in one-time training. Qualitative and quantitative evaluations demonstrate that the proposed PU-EVA outperforms the state-of-the-arts in terms of proximity-to-surface, distribution uniformity, and geometric details preservation.
\end{abstract}

\section{Introduction}
\begin{figure}[t]
	\begin{center}
		\includegraphics[width=1\linewidth]{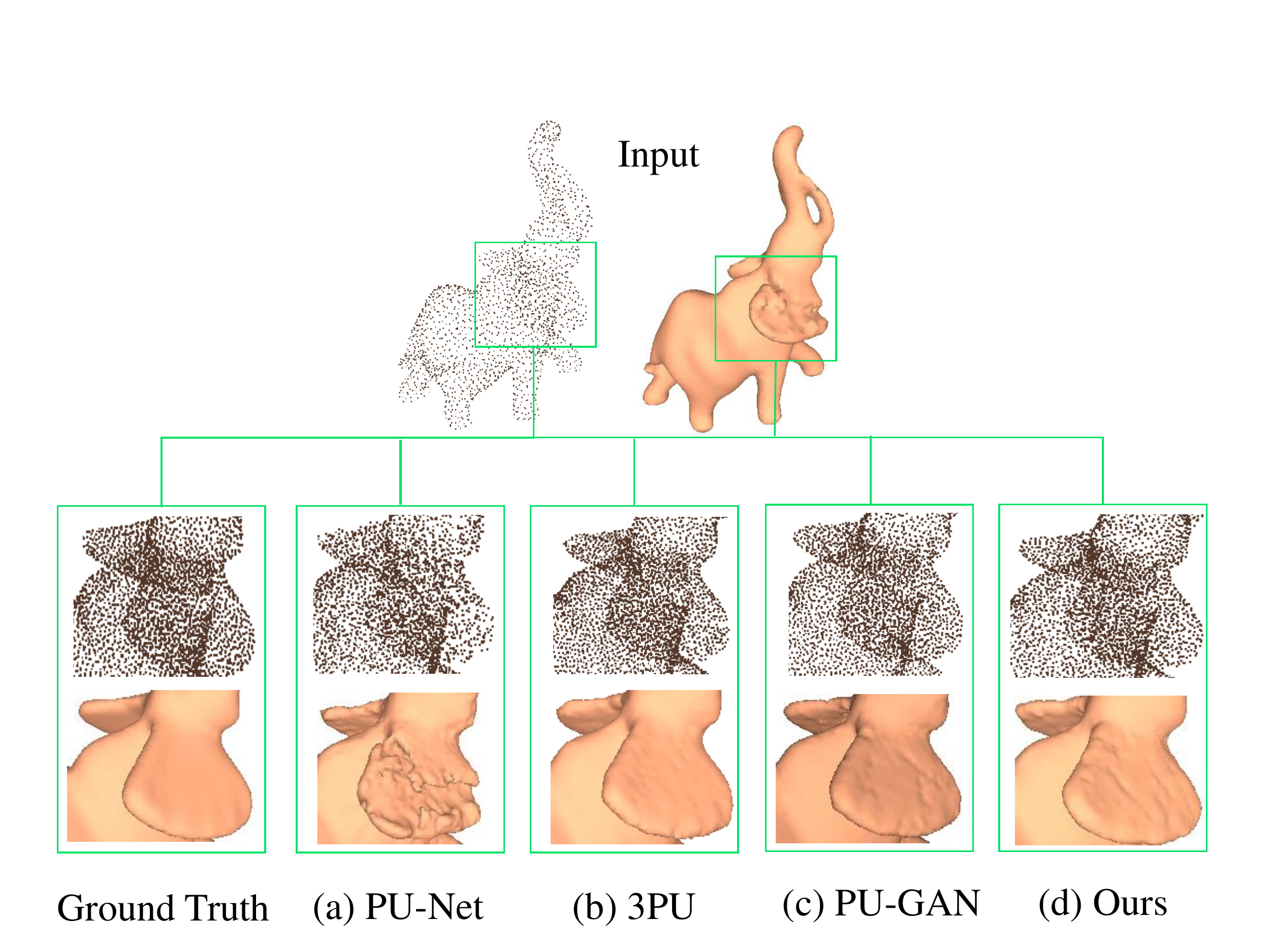}
	\end{center}
	\caption{Point cloud upsampling results of an elephant by using (a) PU-Net~\cite{yu2018pu}, (b) 3PU~\cite{yifan2019patch}, (c) PU-GAN~\cite{li2019pu} and (d) our PU-EVA. Note that our PU-EVA with edge vector based approximation generates points more uniform with fine-grained details, compared to other state-of-the-arts.}
	\label{fig:show}
\end{figure}

With increasing capability of acquiring point clouds as a direct approximation of an object or scene surface from 3D scanning sensors, typical applications processing raw points are prevailing, such as point-based semantic understanding~\cite{armeni20163d,he2019geonet,su2015multi,tang2020improving,wang2019deep}, point clouds rendering~\cite{atienza2019conditional,javaheri2020point} and surface reconstruction~\cite{mostegel2017scalable,williams2019deep}, etc. Hence, the quality of input points is critical for the digital designs. However, a couple of varying artifacts, including occlusion, light reflection, surface materials, sensor resolutions and viewing angles, hinder high-quality practical acquisition of point clouds. It is desirable to upsample raw point clouds for a better description of an underlying surface, by converting the sparse, noisy, and non-uniform points into the dense, regular and uniform ones.

The success of image super-resolution~\cite{ledig2017photo,milanfar2017super,shi2016real,van2006image,yang2010image} in 2D computer vision inspires the corresponding techniques for point clouds upsampling. Nevertheless, the intrinsic irregularity of point clouds makes effective implementations for image interpolations not applicable in 3D world. Based on this, different criteria should also be considered for point clouds upsampling. First, the upsampled points should lie on or be close to the underlying surfaces of target objects. Next, the upsampled results should distribute uniformly, instead of being clustered together. Last, the generated points should preserve geometric details of the target objects.

The previous works often upsample points by simply duplicating point features and disentangling them in positions via a set of MLPs~\cite{li2019pu,yifan2019patch,yu2018pu}. The strategy may lead the generated points cluster around original ones and loss geometric details of the target objects. 
To overcome these issues, we are motivated to develop an upsampling strategy by considering the geometric information of the underlying surface. To capture veracious neighboring connectivity, a similarity matrix constructed by edge vectors is devised.
Here, the edges depict the connections between a center point to the neighboring points, by following the concept of EdgeConv~\cite{wang2019dynamic}. The edge vectors are the oriented edges pointing from the center point, encoding features from both of the connecting points.
The significance of edge vectors to the upsampling point are endowed in the similarity matrix, and an edge vector based affine combination is formed to implement their contributions to upsampling. 
According to the linear approximation theorem originally proposed by~\cite{qian2020deep}, the approximation error of affine combinations in local neighborhood can be formulated as a high-order term mathematically. By applying the theorem in our case, a second-order term of Taylor's Expansion is derived as the approximation error of the proposed method, which encourages the data-driven upsampling network converging under such an approximation.
Moreover, the upsampling unit is designed to learn all possible neighboring connectivity from the edge vectors, making the upsampling rate flexible in one-time training. To implement the motivations, we propose a novel design of Edge Vector based Approximation for Flexible-scale Point clouds Upsampling (PU-EVA). To sum up, the contributions of the proposed PU-EVA are listed as below:
\begin{enumerate}
	\item The proposed PU-EVA achieves a flexible upsampling scale in one-time end-to-end training, making the upsampling rate decoupled with the network architecture;
	\item The edge vector based approximation upsample new points by endowing geometric information of the target objects, which benefits the upsampling performance in sharp regions and areas with fine-grained details; 
	\item The proposed edge vector based affine combinations restrict the approximation error within a second-order term of Taylor's Expansion, encouraging the data-driven network converging under such an approximation.
\end{enumerate}

\section{Related work}\label{rela}
\subsection{Optimization-based upsampling methods}
One of the earliest optimization-based methods generates a set of triangles for the sample points by using three-dimensional Voronoi diagram and Delaunay triangulation~\cite{amenta1998new}. 
A parametric-free method of point resampling and surface approximation is developed by using a locally optimal projection operator (LOP) in~\cite{lipman2007parameterization}. 
However, the LOP-based methods require surface smoothness, giving rise to performance struggling around sharp edges and corners.
To this end, an edge-aware point clouds resampling (EAR) method uses implicit models to preserve sharp features, by resampling points away from edges and upsampling them progressively to approach the edge singularities~\cite{huang2009consolidation}, whereas the performance of EAR heavily depends on the given normal values and parameter tuning.
For point clouds completion and consolidation, a new representation of associating surface points with inner points to reside on the extracted meso-skeletons is proposed, and the optimization is conducted under global geometry constraints from the meso-skeletons~\cite{wu2015deep}. The method recovers regions with holes successfully though, it is sensitive to noise and outliers. Overall, the piece-wise smoothness assumption of optimization-based upsampling methods makes the fine-grained patterns missing, causing the performance of this category of methods limited.

\subsection{Learning-based upsampling methods}
Inspired by data-driven approaches and their promising results, taking advantages of deep learning techniques to model the complex geometries has been brought to attention thus far. 
PU-Net~\cite{yu2018pu}, as the first learning-based point upsampling network, learns geometry semantics of point patches based on the framework of PointNet++~\cite{qi2017pointnet++}, and expands the learned multi-scale features to upsample a given point cloud. 
EC-Net~\cite{yu2018ec} utilizes an edge-aware technique to process point clouds and simultaneously recovers 3D point coordinates and point-to-edge distances by a joint loss. 
3PU~\cite{yifan2019patch} proposes a progressive upsampling network using different number of neighbors in subsequent upsampling units. 
PU-GAN~\cite{li2019pu} is a generative adversarial network (GAN)~\cite{goodfellow2014generative} based point cloud upsampling network. It focuses on boosting the quantitative performance of upsampled point clouds by constructing an up-down-up expansion unit, a self-attention unit as well as a compound loss. 
PUGeo-Net~\cite{qian2020pugeo}, as the first geometry-centric network, proposes to upsample points by learning the first and second fundamental forms of the local geometry under the supervision of normals. CAD-PU~\cite{lin2020cad} proposes geometrically intuitive surrogates to improve the quality of the surface approximation for upsampled point clouds, by realizing the curvature-adaptive feature expansion.
Dis-PU~\cite{li2021point} disentangles the upsampling task by formulating two cascaded sub-networks, a dense generator and a spatial refiner. The dense generator infers a coarse output describing the underlying surface roughly, and the spatial refiner adjusts the point locations via the predicted offsets for a fine upsampling result.
PU-GCN~\cite{Qian_2021_CVPR} proposes a Graph Convolutional Networks (GCN) based network by combining a feature extraction module Inception DenseGCN to capture multi-scale information and a point cloud upsampling module NodeShuffle to encode local point information. The proposed NodeShuffle can also be incorporated into other upsampling pipelines for performance improving.

One of the main limitations of these learning-based upsampling methods is that the upsampling rate is fixed during each training, limiting their applications to the real-world upsampling tasks. MAPU-Net~\cite{qian2020deep} handles an arbitrary upsampling factor in one-time training for the first time, by learning unified and sorted interpolation weights and normal-guided displacements in terms of the proposed linear approximation theorem. A self-attention-based refinement module is further devised to circumvent the normal estimation, driving the upsampling results to approach the underlying surface~\cite{qian2021deep}. Meta-PU~\cite{ye2021meta} also supports the arbitrary scales upsampling by using meta learning to predict the weights of the network and change behavior for each scale factor dynamically.
Our proposed upsampling strategy decouples the upsampling rate with network architecture, with an edge vector based approximation solution to generate new points by encoding neighboring connectivity, achieving the flexible upsampling rates in one-time training.
\begin{figure}[t]
	\begin{center}
		\includegraphics[width=1\linewidth]{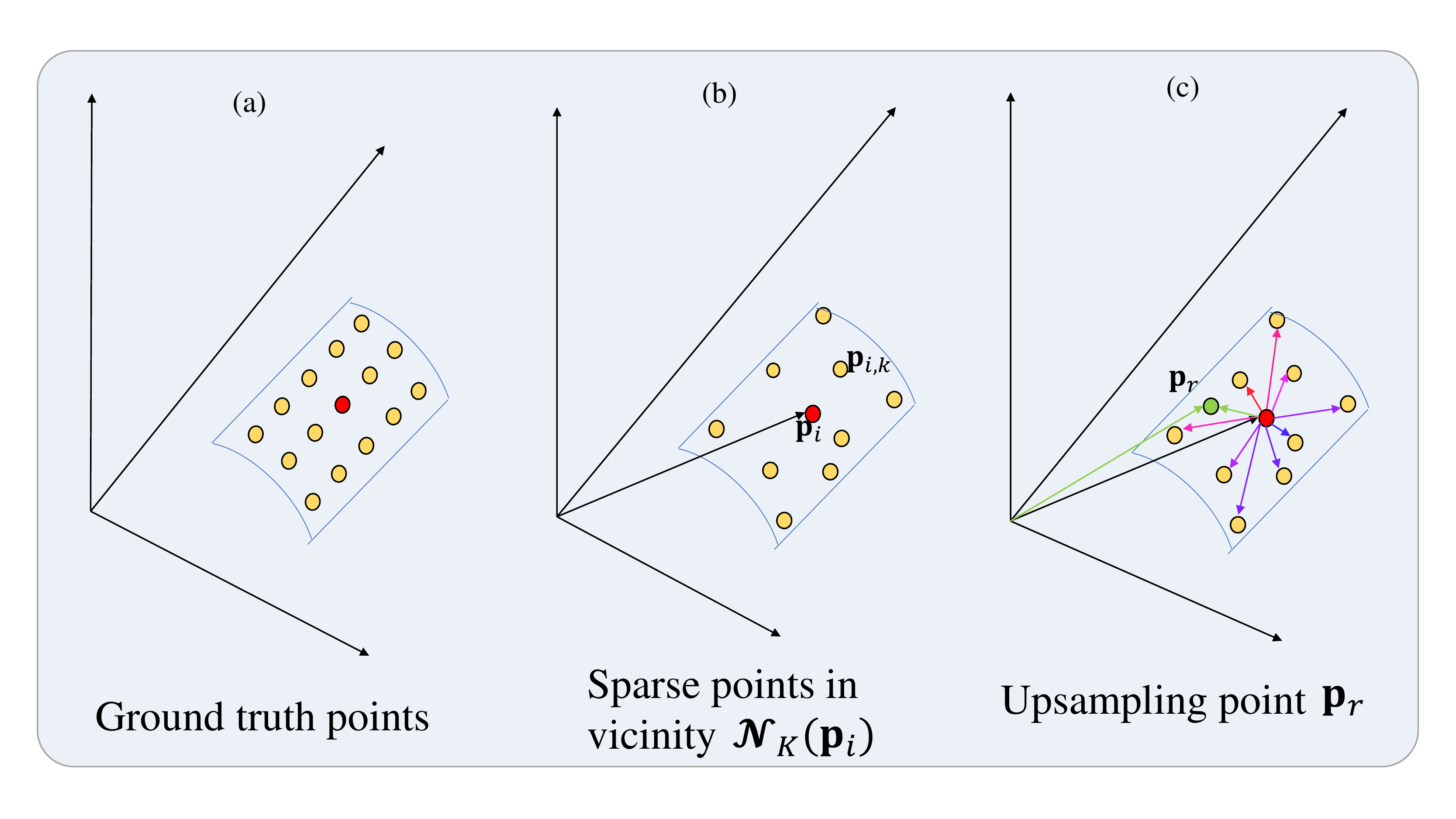}
	\end{center}
	\caption{Edge vector based approximation for point clouds upsampling. (a) Ground truth points. (b) Input sparse points in the vicinity $\mathcal{N}_{K}\left(p_{i}\right)$, centered at point $\mathbf{p}_i$ (red) and surrounded by $K$ nearest neighbors $\mathbf{p}_{i,k}$ (yellow). (c) Points upsampling via edge vector based approximation. Arrows of different colors represent the edge vectors with different significance to the upsampled point $\mathbf{p}_{r}$ (green).}
	\label{fig:math}
\end{figure}

\section{Proposed method}\label{meth}
\begin{figure*}
	\begin{center}
		\includegraphics[width=.85\linewidth]{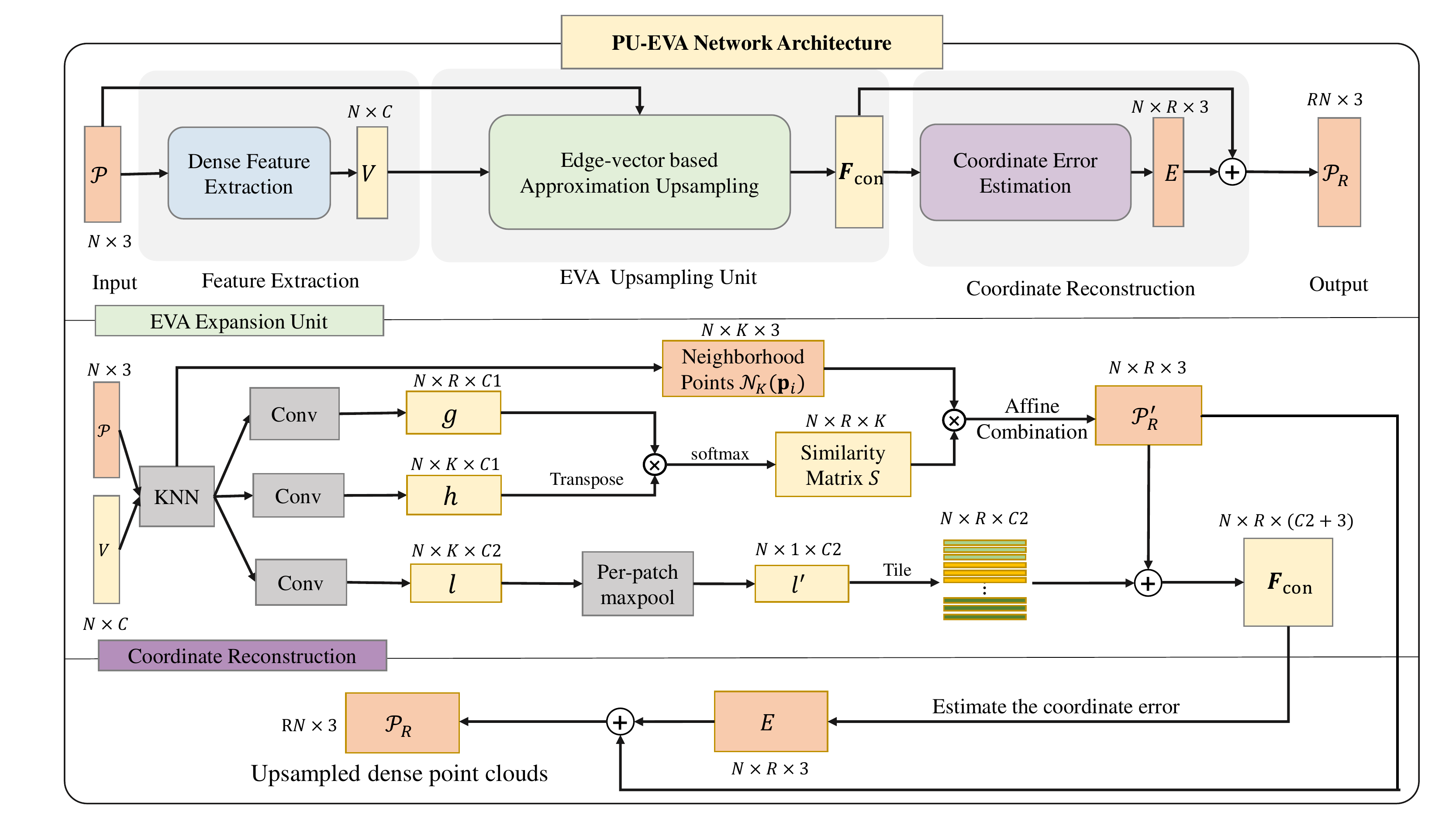}
	\end{center}
	\caption{Architecture of PU-EVA network, consisting of three components: feature extraction, point upsampling and coordinate reconstruction. It takes spare input ($N$ points) and produces dense point clouds ($RN$ points) by a carefully designed edge vector based approximation upsampling method, where $R$ is the upsampling rate.}
	\label{fig:arch}
\end{figure*}
In this section, we present an end-to-end data-driven framework for point clouds upsampling, dubbed Edge Vector based Approximation for Flexible-scale Point clouds Up-sampling (PU-EVA). As shown in Figure~\ref{fig:arch}, the proposed PU-EVA consists three elements: dense feature extraction, edge vector based approximation upsampling, and coordinate reconstruction. The detailed architecture is explained below.

\subsection{Dense feature extraction}~\label{4.2}
Given an input sparse point cloud $\mathcal{P}=\left\{\mathbf{p}_{i} \in \mathbb{R}^{3} | (p_{i}^{x}, p_{i}^{y}, p_{i}^{z}) \right\}_{i=1}^{N}$, where $N$ is the number of points. The objective is to generate dense points $\mathcal{P}_R=\left\{\mathbf{p}_{r} \in \mathbb{R}^{3} | (p_{r}^{x}, p_{r}^{y}, p_{r}^{z})\right\}_{r=1}^{RN}$ approximating the underlying surface, where $R$ is the upsampling rate\footnote{In our formulation, the cursive letter $\mathcal{P}$ represents a point cloud, and the bold letter $\mathbf{p}$ represents the point coordinates.}. As show in Figure~\ref{fig:math} (b), consider a small vicinity $\mathcal{N}_K(\mathbf{p}_i) \subseteq \mathcal{P}$, the red point denotes the center point $\mathbf{p}_i$, and the yellow points indicate $K$ nearest neighbors $\mathbf{p}_{i,k}$.

In dense feature extraction, a dense dynamic feature extraction method in~\cite{yifan2019patch} is applied on each vicinity $\mathcal{N}_K(\mathbf{p}_i) \subseteq \mathcal{P}$, by adopting the EdgeConv~\cite{wang2019dynamic} as a basic block. In essence, four dense blocks are employed for extraction. To leverage features of different layers, skip-connections are applied within and between the dense blocks. Within the dense block, three convolutional layers are skip-connected, the output features are passed to all subsequent layers, and a max-pooling layer is employed at the end for input order invariant; between the dense blocks, the features produced by each block are fed to all following blocks. In this way, the point feature $V=\left\{\mathbf{v}_i\right\}_{i=1}^{N}$ of $N \times C$ are extracted from the sparse input $\mathcal{P}$, where $C$ stands for the number of dimensions. The enriched geometric features with long-range and non-local information from multiple scales are embedded in the obtained point feature $V$ for further edge vector based operations.

\subsection{Edge vector based approximation upsampling }~\label{4.3}
The Edge Vector based Approximation (EVA) is put forward in the vicinity $\mathcal{N}_K(\mathbf{p}_i)$ for upsampling, as illustrated in Figure~\ref{fig:math} (c). A similarity matrix $S$ determines the significance of edge vectors to the upsampling point. Specifically, the edge vector $\Delta \mathcal{F}_{i,k}$ with respect to $K$ neighbor points are defined as,
\begin{equation}
\begin{aligned}
\Delta  \mathcal{F}_{i,k} =
& (\mathbf{v}_{i,k} -  \mathbf{v}_{i}) \oplus
\mathbf{v}_{i,k} \oplus
\mathbf{v}_i\\
& \oplus (\mathbf{p}_{i,k}-\mathbf{p}_i) \oplus
\mathbf{p}_{i,k} \oplus
\mathbf{p}_i \oplus
d_{i,k},
\end{aligned}
\end{equation}
\noindent where $\mathbf{p}_i$ and $\mathbf{v}_i$ are point coordinates and point features of the center point, $\mathbf{p}_{i,k}$ and $\mathbf{v}_{i,k}$ are point coordinates and point features of $K$ neighbor points, and $d_{i,k}$ is the Euclidean distance between the center point and the neighbors. Similarly, the edge vector $\Delta \mathcal{F}_{i,r}$ is obtained based on $R$ anchor points. Here, $R$ anchors points are selected from $K$ neighbor points, implying that $R$ is no bigger than $K$. The correlations between the edge vectors $\Delta \mathcal{F}_{i,k}$ and $\Delta \mathcal{F}_{i,r}$ form the elements of similarity matrix $S$ with dimension $N \times R \times K$.

As shown in Figure~\ref{fig:arch}, the 1$\times$1 convolutional operators $\phi(\cdot)$ and $\psi(\cdot)$ (described as ``Conv") map the edge vectors into high dimensional feature space as,
\begin{equation}
g = \{\phi_1 \left(\Delta \mathcal{F}_{i, r}\right)| g \in  \mathbb{R}^{N\times R\times C_1}\},
\end{equation}
\begin{equation}
h = \{\phi_2 \left(\Delta \mathcal{F}_{i, k}\right)| h \in  \mathbb{R}^{N\times K\times C_1}\}, 
\end{equation}
\begin{equation}
l = \{\psi \left(\Delta \mathcal{F}_{i, k}\right)| l \in  \mathbb{R}^{N\times K\times C_2}\}. 
\end{equation}

\noindent where $C_1$ and $C_2$ are in different number of dimensions. The correlations among edge vectors are attained via a softmax function as,
\begin{equation}
w_{k} = f_{\text{softmax}} (g \cdot h^{\top}), \quad \text{s.t.} \sum_{k=1}^{K} w_{k}=1, \forall w_{k} > 0,
\label{eq:soft}
\end{equation}
\noindent where $w_{k}$ normalized by the softmax function $f(\cdot)$ encodes the significance of edge vectors to the upsampled point $\mathbf{p}_{r}$ by computing the local correlations, as demonstrated in different color of arrows in Figure~\ref{fig:math} (c). 

To endow all possible neighboring connectivity from the edge vectors for upsampling, the $R$ anchor points are selected randomly from $K$ neighbor points in each iteration. Although $R$ is smaller than $K$, the random selection guarantees the overall information in the local neighborhood are learned by the network during iteration. The number of generated points can be set as $R$ times of the original points in the inference stage. As a result, the dimension of the similarity matrix $N\times R\times K$ depends on the local selection only, which decouples with the network architecture. The upsampling rate $R$ is an arbitrary positive integer smaller than $K$. Furthermore, the local feature $l$ mapped by convolution $\psi(\cdot)$ based on the edge vector $\Delta \mathcal{F}_{i,r}$ is max-pooled and tiled up as $l^{\prime}$ to preserve rich local geometric patterns, which encourages new points describing sharp regions and areas with fine-grained details better.

By obtaining the similarity matrix $S$, the correlations among edge vectors can be considered as the coefficients $w_{k}$ of point based affine combinations,
\begin{equation}
\sum_{k=1}^{K} w_{k} \mathbf{p}_{i,k}=\sum_{k=1}^{K} w_{k} (p_{i,k}^{x}, p_{i,k}^{y}, p_{i,k}^{z}).\label{eq:aff}
\end{equation}
The edge vector based affine combinations of local neighborhood approximate the underlying surface across the target object. Theoretically, the approximation error can be accumulated across the curved surface iteratively. Referring to the linear approximation theorem~\cite{qian2020deep}, the error can be expressed explicitly as a high-order term. Based on this, we derive the approximation error of the proposed edge vector based affine combination as a second-order term of Taylor's Expansion. The error estimation encourages the data-driven network converging under such an approximation. The mathematical reasoning is elaborated below.

Assume that the vicinity $\mathcal{N}_K(\mathbf{p}_i)$ is local smoothness at point $\mathbf{p}_i$. According to the implicit function theorem~\cite{hildebrand1962advanced}, the vicinity satisfies $f\left(\mathbf{p}_{i,k}\right)=0, \forall \mathbf{p}_{i,k} \in \mathcal{N}_K( \mathbf{p}_i)$, where $f(\cdot)$ is a smooth implicit function. If the partial derivative $\frac{\partial {f}}{\partial {p^z}}$ exists, the local curved surface can be expressed explicitly by a height function $F: \mathbb{R}^{2} \rightarrow \mathbb{R}: {p^z}=F(p^{x}, p^{y})$. Theoretically, given $p_{r}^{x}$ and $p_{r}^{y}$, $p_{r}^{z}$ on the curved surface can be obtained via the height function $F(\cdot)$. 

However, the analytical solution on a curved surface is too computational expensive to acquire. The corresponding numerical solution can be approximated via Taylor's Expansion at any point $\mathbf{p}_j$ in local neighborhood as, 
\begin{equation}
\begin{split}
{p^z}&=F\left(p_{j}^{x}, p_{j}^{y}\right)+\nabla F\left(p_{j}^{x}, p_{j}^{y}\right)^{\top} \cdot\left(p^{x}-p^{x}_{j}, p^{y}-p^{y}_{j}\right)\\
&+O(\left(p^{x}-p^{x}_{j},p^{y}-p^{y}_{j}\right)^{2}).
\label{eq:tay}
\end{split}
\end{equation}

A linear function $F_a(\cdot)$ can be defined as,
\begin{equation}
F_a(p^x, p^y) \triangleq F\left(p_{j}^{x}, p_{j}^{y}\right)
+\nabla F\left(p_{j}^{x}, p_{j}^{y}\right)^{\top} \cdot\left(p^{x}-p^{x}_{j}, p^{y}-p^{y}_{j}\right).
\label{eq:linear}
\end{equation}

As a result, Eq.~\ref{eq:tay} can be expressed as,
\begin{equation}
\begin{split}
{p^z} &= F_a(p^x, p^y) + O(\Delta ^2) \approx F_a(p^x, p^y).
\label{eq:tay2}
\end{split}
\end{equation}

\noindent ${p^z}$ on the curved surface is approximated by the corresponding point on the tangent plane. Considering the involved gradients calculation is complex, referring to~\cite{qian2020deep}, any point can be constructed by the existing points $p_{i,k}^{x}$ and $p_{i,k}^{y}$ in the two-dimensional local region as,
\begin{equation}
(p_{r}^{x}, p_{r}^{y}) = \sum_{k=1}^{K} w_{k} \ (p_{i,k}^{x}, p_{i,k}^{y}),
\label{eq:close}
\end{equation}
\noindent with non-negative coefficients $w_{k}$. By substituting Eq.~\ref{eq:close} into Eq.~\ref{eq:tay2}, $p_{r}^{z}$ on the curved surface is further approximated as,
\begin{equation}
\begin{aligned}
{p_r^z} &\approx F_a(p_r^x, p_r^y) =
F_a \left(\sum_{k=1}^{K} w_{k} \ (p_{i,k}^{x}, p_{i,k}^{y}) \right).
\end{aligned}
\end{equation}
Since $\mathbf{p}_{j}$ is a given point in local neighborhood,  $F\left(p_{j}^{x}, p_{j}^{y}\right)$ and $\nabla F\left(p_{j}^{x}, p_{j}^{y}\right)^{\top}$ in Eq. \ref{eq:linear} can be treated as the constants, thus, $p_{r}^{z}$ can be further approximated as,
\begin{equation}
\begin{aligned}
F_a \left(\sum_{k=1}^{K} w_{k} \ (p_{i,k}^{x}, p_{i,k}^{y}) \right) =\sum_{k=1}^{K} w_{k} F_a(p_{i,k}^{x}, p_{i,k}^{y})
\approx \sum_{k=1}^{K} w_{k} {p_{i,k}^z}.
\end{aligned}
\end{equation}

To this end, the generated points $\mathbf{p}_{r}$ residing on the curved surface are approximated by the edge vector based affine combination as,
\begin{equation}
\begin{split}
(p_{r}^{x}, p_{r}^{y}, p_{r}^{z})
&\approx (p_{r}^{x}, p_{r}^{y}, \sum_{k=1}^{K} w_{k}  p_{i,k}^{z}),\\
&=  (\sum_{k=1}^{K} w_{k} (p_{i,k}^{x}, p_{i,k}^{y}), \sum_{k=1}^{K} w_{k}  p_{i,k}^{z}),\\
&=\sum_{k=1}^{K} w_{k} (p_{i,k}^{x}, p_{i,k}^{y}, p_{i,k}^{z}).
\end{split}
\end{equation}
\noindent where the edge vectors are constructed by the given sparse points, and the approximation error is restricted within a second-order error term of Taylor's Expansion $O(\Delta ^2)$.

Overall, the approximated $\mathcal{P}_R^{\prime}$ can be written as,
\begin{equation}
\mathbf{p}_{r}^{\prime} = \sum_{k=1}^{K} w_{k} \mathbf{p}_{i,k}, \quad \text{s.t.} \sum_{k=1}^{K} w_{k}=1, \forall w_{k} > 0,
\end{equation}
\noindent with the coefficients decided by Eq.~\ref{eq:soft}.

\subsection{Coordinate reconstruction}~\label{4.4}
The output of edge vector based affine combinations concatenates with the expanded feature $l^{\prime}$, yielding a concatenated feature $\mathbf{F}_\text{con}$ for coordinate reconstruction, as demonstrated in the bottom of Figure~\ref{fig:arch},
\begin{equation}
\mathbf{F}_\text{con} = \mathcal{P}_R^{\prime} \oplus \textit{Tile} (l^{\prime}).
\end{equation}

Under the guidance of concatenated feature $\mathbf{F}_\text{con}$ enriched by the expanded local feature $l^{\prime}$ with more information from sharp regions and areas with fine-grained details, the coordinate error is estimated by a series of MLPs. As explained in Section~\ref{4.3}, the approximation error can be considered as the second-order term of Taylor's Expansion $O(\Delta ^2)$, which encourages the upsampling network converging. Thereby, the desirable parameters of upsampling network are obtained as long as the supervision signals are fidelity to the ground truth. The coordinate error estimation network $E$ utilizes MLPs under the supervision of ground truth $\mathcal{Q}=\left\{\mathbf{q}_{i} \in \mathbb{R}^{3} | (q_{i}^{x}, q_{i}^{y}, q_{i}^{z}) \right\}_{i=1}^{RN}$, to regress the approximated $\mathcal{P}_R^{\prime}$ towards the final output coordinates of point clouds $\mathcal{P}_R$, with the involved loss functions articulated below.

\textbf{Modified Chamfer Distance (CD) loss} in ~\cite{achlioptas2018learning,yifan2019patch} measures the similarity between the upsampled point clouds $\mathcal{P}_{R}$ and the ground truth $\mathcal{Q}$, which is,
\begin{equation}
\begin{split}
\mathcal{L}_\text{CD}(\mathcal{P}_{R},\mathcal{Q})
&=\frac{1}{|\mathcal{P}_{R}|}\sum_{p\in \mathcal{P}_{R}}^{} \xi (\underset{q\in \mathcal{Q}}{min}\left \| p-q \right \|_{2}^{2}) \\
&+\frac{1}{|\mathcal{Q}|}\sum_{q\in \mathcal{Q}}^{} \xi ( \underset{p\in \mathcal{P}_{R}}{min}\left \| q-p\right \|_{2}^{2}),
\end{split}
\end{equation}
\noindent where the operator $\|\cdot\|_{2}^{2}$ stands for the squared Euclidean norm, and function $\xi(d)= \begin{cases}d, & d \leq \delta, \\ 0, & \text {otherwise,}\end{cases}$ filters outliers above a specified threshold. CD loss facilitates the produced points lying on or being closer to the underlying surface.

\textbf{Uniform loss} adopted in~\cite{li2019pu} are defined as,
\begin{equation}
	\mathcal{L}_\text{uni}=\sum_{j=1}^{M} U_\text{imbalance}(S_{j})\cdot U_\text{clutter}(S_{j}),
\end{equation}
\noindent where $U_\text{imbalance}$ considers the non-local distribution uniformity by evaluating the deviation of $\left | S_{j} \right |$ from $\hat{n}$,
\begin{equation}
U_\text{imbalance}(S_{j})=\frac{(\left | S_{j} \right |-\hat{n})^{2}}{\hat{n}}.
\end{equation}
\noindent The point subsets $\{S_{j}| j=1\dots M\}$ are cropped by a ball query of radius $r_d$ in seed points, and the seed points are picked by Farthest Point Sampling (FPS) from $\mathcal{P}_R$. $\hat{n}$ is the expected number of points in $S_{j}$. $U_\text{clutter}$ takes account of the local uniformity as,
\begin{equation}
U_\text{clutter}(S_{j})=\sum_{k=1}^{\left | S_{j} \right |}\frac{( d_{w,k}-\hat{d} )^{2}}{\hat{d}},
\end{equation}
\noindent $d_{w,k}$ indicates the distance between each point $\mathbf{p}_{w}$ in $S_{j}$ and its nearest neighbor $\mathbf{p}_{w,k}$, and $\hat{d}$ is the expected point-to-neighbor distance.

The proposed PU-EVA network is trained by minimizing the joint loss function in an end-to-end fashion:
\begin{equation}
\mathcal{L}(\theta) = \alpha \mathcal{L}_\text{CD} +   \beta\mathcal{L}_\text{uni} +\gamma \left \| \theta \right \|^{2},
\label{eq:loss}
\end{equation}
\noindent where $\theta$ denotes the parameters in our PU-EVA network, $\alpha$ and $\beta$ balance the CD loss and the uniform loss,  and $\gamma$ indicates the multiplier of weight decay.

\section{Experiments}\label{exp}
\begin{table*}
	\begin{center}
		\label{tab:proxi}
		\scalebox{0.9}{
			\begin{tabular}{|l|ccc|ccc|ccc|}
				\hline
				\bf Methods & \multicolumn{3}{c|}{\bf Sparse (256) input} & \multicolumn{3}{c|}{\bf Medium (2,048) input} & \multicolumn{3}{c|}{\bf{Dense (4,096) input}} \\
				\bf $\mathbf{(10^{-3})}$ &\bf CD & \bf HD & \bf P2F & \bf CD & \bf HD & \bf P2F & \bf CD & \bf HD & \bf P2F \\
				\hline\hline
				EAR~\cite{huang2013edge} &-&-&-& 0.520 & 7.370 & 5.820 &-&-&-\\
				PU-Net~\cite{yu2018pu} & 2.461 & 15.370 & 13.099 & 0.720 & 8.940 & 6.840 & 0.247 & 2.802 & 12.033 \\
				3PU~\cite{yifan2019patch} & 2.177 & \bf 12.672 & 10.328 & 0.490 & 6.110 & 3.960 & 0.446 & 4.225 & 4.281 \\
				PU-GAN~\cite{li2019pu} & 2.072 & 16.592 & \bf 8.055 & 0.280 & 4.640 & \bf 2.330 & 0.131 & \bf 1.284 & 1.687 \\
				\bf Ours & \bf 1.784 & 13.939 & 8.727 & \bf 0.266 & \bf 3.070 & 2.362 & \bf 0.123 & 1.394 & \bf 1.416\\
				\hline
		\end{tabular}}
	\end{center}
	\caption{Quantitative comparison of proximity-to-surface with various input point resolutions.}
\end{table*}

\begin{figure*}
	\begin{center}
		\includegraphics[width=1\linewidth]{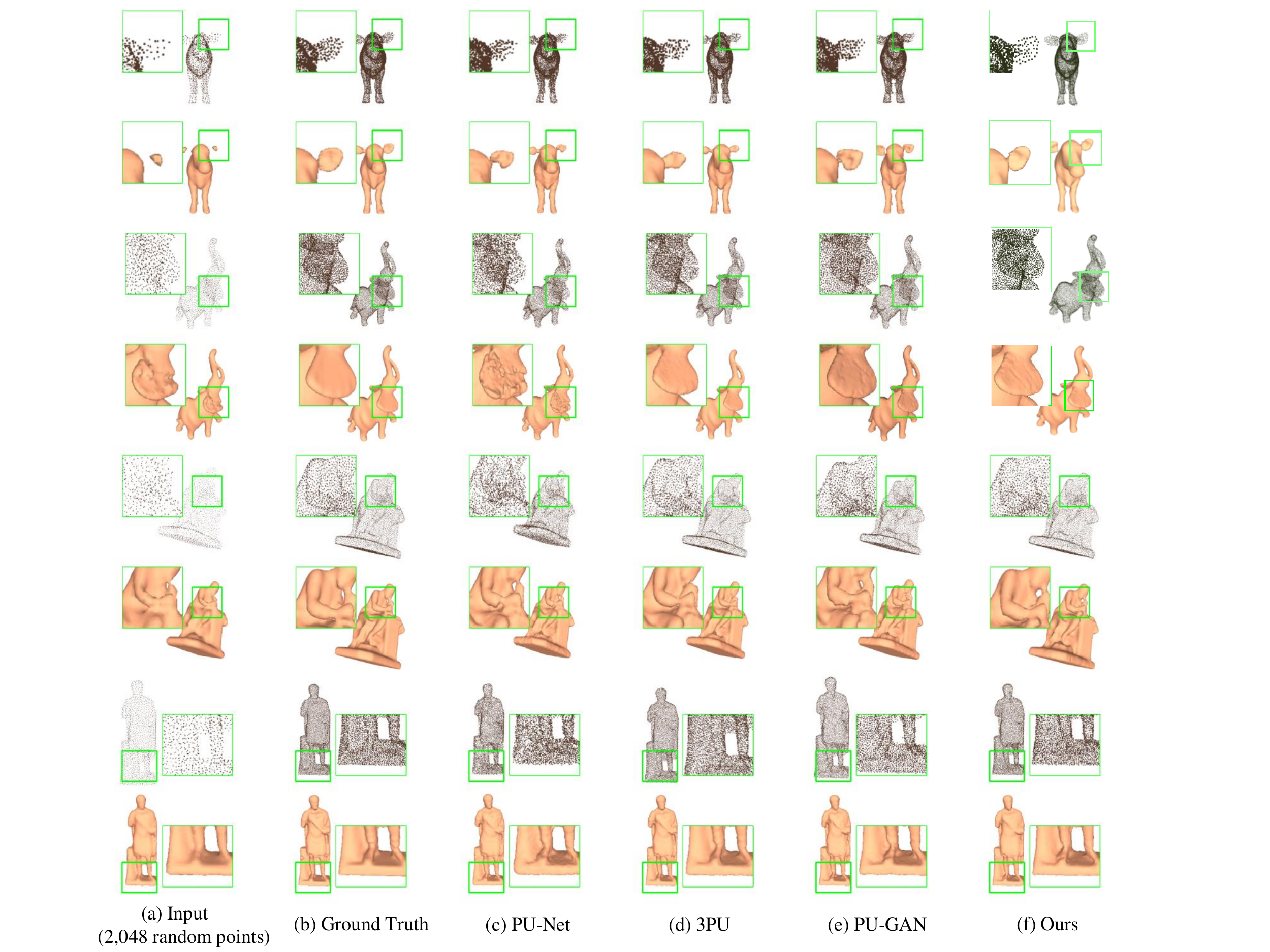}
	\end{center}
	\caption{Comparison of point clouds upsampling ($4 \times$) and surface reconstruction with state-of-the-art methods (c-f) from inputs (a).}
	\label{fig:up_visual}
\end{figure*}

\begin{table}
	\begin{center}
		\label{tab:unif}
		\scalebox{0.9}{
			\begin{tabular}{|l|ccccc|}
				\hline
				\bf Methods & \multicolumn{5}{c|}{\bf Uniformity for different $p$} \\
				\bf $\mathbf{(10^{-3})}$ & \bf 0.4\% & \bf 0.6\% & \bf 0.8\% & \bf 1.0\% & \bf 1.2\% \\
				\hline\hline
				EAR~\cite{huang2013edge} & 16.84 & 20.27 & 23.98 & 26.15 & 29.18 \\
				PU-Net~\cite{yu2018pu} & 29.74 & 31.33 & 33.86 & 36.94 & 40.43 \\
				3PU~\cite{yifan2019patch} & 7.51 & 7.41 & 8.35 & 9.62 & 11.13 \\
				PU-GAN~\cite{li2019pu} & 3.38 & 3.49 & 3.44 & 3.91 & 4.64\\
				\bf Ours & \bf 2.26 & \bf 2.10 & \bf 2.51 & \bf 3.16 & \bf 3.94 \\
				\hline
		\end{tabular}}
	\end{center}
	\caption{Quantitative comparison of uniformity for different $p$.}
\end{table}

In this section, several experiments are conducted to compare our method with state-of-the-art point upsampling methods quantitatively and qualitatively, and various aspects of our model are evaluated.
\begin{figure*}[t]
	\begin{center}
		\includegraphics[width=0.8\linewidth]{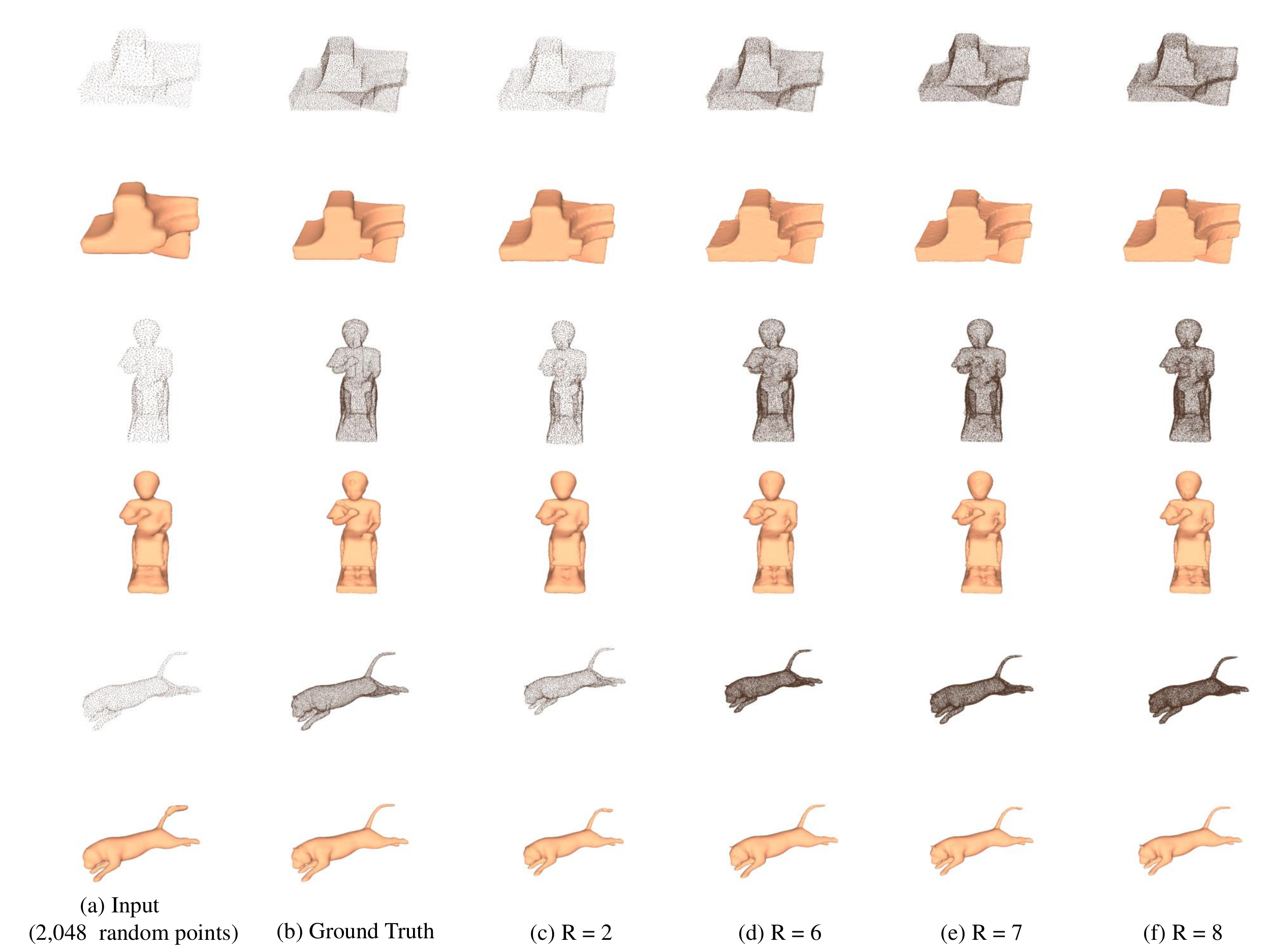}
	\end{center}
	\caption{Point clouds upsampling of PU-EVA with various upsampling rate.}
	\label{fig:diff_r}
\end{figure*}

\subsection{Experimental setup}
\subsubsection{Data preparation}
We train all these upsampling methods based on a benchmark dataset mentioned in~\cite{li2019pu} for fair comparison, where 147 objects covering a rich variety of shapes are collected. By following~\cite{li2019pu}, the same 120 objects are used for training, and the remaining are used for testing. For quantitative comparison, the referenced ground truth point distribution $\mathcal{Q}$ of 8,192 points on the patch is sampled from original meshes by Poisson disk sampling~\cite{corsini2012efficient}. In the training stage, the input spare point clouds $\mathcal{P}$ are randomly selected from $\mathcal{Q}$ on-the-fly. In the testing stage, $6 \times$ points are randomly selected from ground truth first, then $4 \times$ points are further sampled by FPS~\cite{eldar1997farthest}. Data augmentation techniques are also applied to rotate, shift and scale data randomly to avoid over-fitting. Throughout the experiments, we train PU-EVA for 300 epochs with a mini batch size 32 on GPU NVIDIA 2080Ti. All models converge before the maximum epochs.

\subsubsection{Network configurations}
In the dense feature extraction unit, four dense blocks based on EdgeConv are employed. To leverage the features of different layers, skip-connections are applied within and between the dense blocks. Concretely, within the dense block, three convolutional layers of output channels 24 are skip-connected, the output features are passed to all subsequent layers, and a max-pooling layer is employed at the end; between the dense blocks, the features produced by each block are fed to all following blocks, resulting in the output channels of 120, 240, 360, and 480, sequentially.

In the EVA upsampling unit, $K$ and $R$ are set as 12 and 6 to obtain the neighbor points and the anchor points, empirically. $C_1$ and $C_2$ are set as 240 and 480, respectively. $l$ is max-pooled into dimension of $N\times 1\times 480$ before tiling by the upsampling rate ($R=6$) to expand as $l^{\prime}$ of $N\times 6\times 480$. The output of $\mathcal{P}_R^{\prime}$ $N\times 6\times 3$ is then concatenated with the expanded $l^{\prime}$ to get features $\mathbf{F}_{\mathrm{con}}$ with dimension $N\times 6\times (480+3)$ for coordinates reconstruction. 

In the coordinate reconstruction unit, the error estimation module utilizes three fully connect layers (MLPs) of output channels 256, 128, and 64, followed by the last coordinate regression layer of output dimension $RN \times 3$. All the convolutional layers and fully connected layers in the network are followed by the ReLU activation function, except for the last coordinate regression layer.

\subsubsection{Evaluation metrics}
Four evaluation metrics are employed for quantitative comparison of proximity-to-surface and distribution uniformity. To measure the deviations between upsampled points and ground truth, three commonly used metrics Chamfer Distance (CD), Hausdorff Distance (HD)~\cite{birsan2005one,rockafellar2009variational}, and point-to-surface distance (P2F) are adopted to evaluate the upsampling performance.

For $\mathcal{L}_\text{CD}$, $\delta$ in function $\xi$ is set as 100. For $\mathcal{L}_\text{uni}$, we first use FPS to pick $M$ seed points in $\mathcal{P}_R$ and avails a ball query of radius $r_{d}$ to crop each seed for point subset $\left\{S_{j} \mid j=1 \ldots M\right\}$ to evaluate $\mathcal{P}_R$ during training. $M$ is chosen as 1,000, and the actual mesh of testing model is used to find $S_j$ geodesically instead of cropping $S_j$ by ball query. According to $r_{d}=\sqrt{p}$, $p$ is set as 0.4\%, 0.6\%, 0.8\%, 1.0\%, and 1.2\%, respectively. $\alpha$, $\beta$ and $\gamma$ are set as 150, 10, and 1 empirically. In the implementation,  All these metrics are compared over the whole point clouds, and the smaller values indicate the better upsampling quality.

\subsection{Quantitative and qualitative comparison}
\subsubsection{Quantitative results}
Table 1 summarizes quantitative comparison of the proposed PU-EVA and other state-of-the-arts, i.e., EAR~\cite{huang2013edge}, PU-Net~\cite{yifan2019patch}, 3PU~\cite{yifan2019patch} and PU-GAN~\cite{li2019pu}, in terms of proximity-to-surface. The comparison is based on $4 \times$ upsampling, various input resolutions are evaluated, containing sparse points (256), medium points (2,048), and dense points (4,096), respectively. PU-EVA achieves the best performance with the lowest deviations from surface consistently, which means points generated from our method are closer to the ground truth. Besides, the upsampling rates of PU-Net~\cite{yifan2019patch} and PU-GAN~\cite{li2019pu} are tangled with the network architectures, with features expanded by replication and rearrangement, neglecting the complex geometric information contained in the latent object surface. Although 3PU~\cite{yifan2019patch} shows a competitive capability to deal with flexible upsampling rates, the training process is complicated and more subsets are required for a higher upsampling rate, which is required to be in powers of 2. 

Table 2 reports quantitative comparison of points distribution uniformity for different $p$. We compare with optimization based EAR~\cite{huang2013edge} and three deep learning based methods, i.e., PU-Net~\cite{yifan2019patch}, 3PU~\cite{yifan2019patch} and PU-GAN~\cite{li2019pu}.
The uniformity of our results stays the lowest for all different $p$, indicating that PU-EVA obtains the best distribution uniformity compared to state-of-the-arts over varying scales.
Furthermore, the performance of EAR~\cite{huang2013edge} heavily depends on normal values and parameter tuning.

\subsubsection{Qualitative results}
\begin{table}
	\begin{center}
		\label{tab:gau}
		\scalebox{0.85}{
			\begin{tabular}{|l|cc|cc|cc|}
				\hline
				\bf Methods & \multicolumn{2}{c|}{ $\boldsymbol{\sigma=0}$ } & \multicolumn{2}{c|}{$\boldsymbol{\sigma=0.01}$} & \multicolumn{2}{c|}{$\boldsymbol{\sigma=0.02}$} \\
				\bf $\mathbf{(10^{-3})}$ & \bf CD & \bf HD & \bf CD & \bf HD & \bf CD & \bf HD\\
				\hline\hline
				PU-GAN~\cite{li2019pu} & 0.280 & 4.640 & 0.512 & 6.496 & 0.912 & 11.326\\
				\bf Ours & \bf 0.266 & \bf 3.070 & \bf 0.464 & \bf 5.501 & \bf 0.864 & \bf 9.178 \\
				\hline
		\end{tabular}}
	\end{center}
	\caption{Quantitative comparison of upsampling results with different additive Gaussian noise levels.}
\end{table}

\begin{table}
	\begin{center}
		\label{tab:tylor}
		\scalebox{0.8}{
			\begin{tabular}{|l|cc|cc|cc|}
				\hline
				\bf 2nd-order Error &\multicolumn{2}{c}{\bf Sparse (256)} &
				\multicolumn{2}{|c}{\bf Medium (2,048)} &
				\multicolumn{2}{|c|}{\textbf{Dense (4,096)}} \\
				\bf Term $\mathbf{(10^{-3})}$ & \bf CD & \bf HD & \bf CD & \bf HD & \bf CD & \bf HD\\
				\hline\hline
				without & 2.318 & 18.713 & 0.312 & 4.622 & 0.157 & 5.535 \\
				\bf with & \bf 1.784 & \bf 13.939 & \bf 0.266 & \bf 3.070 & \bf 0.123 & \bf 1.394\\
				\hline
		\end{tabular}}
	\end{center}
	\caption{Quantitative comparison of PU-EVA with and without the second-order error term of Taylor's Expansion from sparse to dense input point resolutions.}
\end{table}

\begin{table*}
	\begin{center}
	    \label{tab:abl}
		\scalebox{0.9}{
			\begin{tabular}{|l|cc|cc|cc|}
				\hline
				\bf Upsampling Unit &\multicolumn{2}{c|}{\bf Sparse (256) input} & \multicolumn{2}{c|}{\bf Medium (2,048) input} & \multicolumn{2}{c|}{\textbf{Dense (4,096) input}} \\
				\bf $\mathbf{(10^{-3})}$ & \bf CD & \bf HD & \bf CD & \bf HD & \bf CD & \bf HD \\
				\hline\hline
				Feature Duplication & 2.94 & 24.67 & 0.38 & 3.99 & 0.13 & 4.47\\
				NodeShuffle~\cite{qian2019pu} & 1.86 & 15.56 & 0.27 & 4.66 & 0.12 & 3.11 \\
				\bf EVA Upsmapling & \bf 1.78 & \bf 13.93 &\bf 0.26 & \bf 3.07 & \bf 0.12 & \bf 1.39 \\
				\hline
		\end{tabular}}
	\end{center}
	\caption{Quantitative comparison of different upsampling units from sparse to dense input point resolutions.}
\end{table*}

\begin{figure}[t]
	\begin{center}
		\includegraphics[width=1\linewidth]{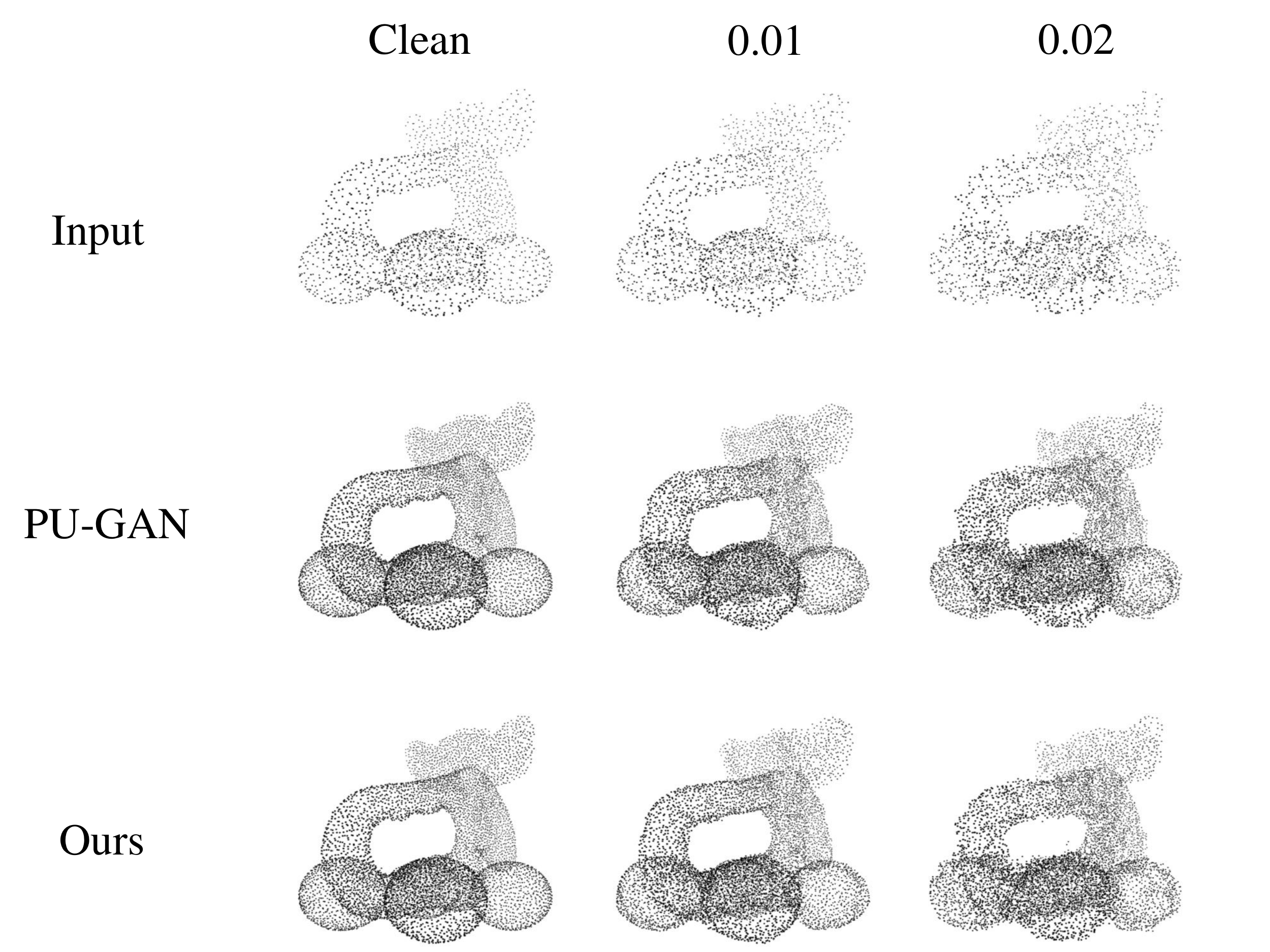}
	\end{center}
	\caption{Qualitative comparison of upsampling results with different additive Gaussian noise levels: 0, 0.01 and 0.02 from left to right, respectively.}
	\label{fig:noise}
\end{figure}

\begin{figure}[t]
	\begin{center}
		\includegraphics[width=1\linewidth]{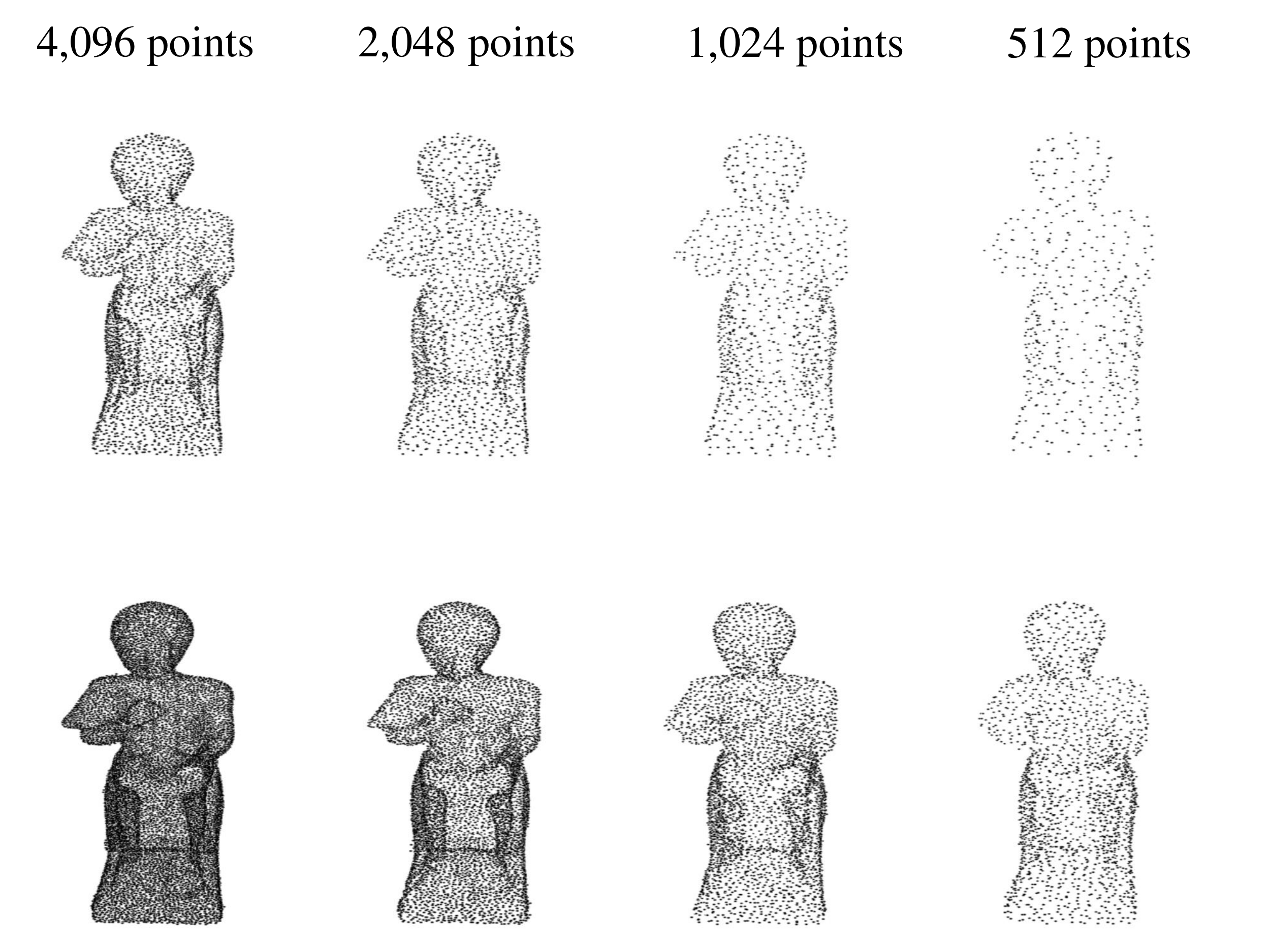}
	\end{center}
	\caption{Upsampling results with varying sizes of input.}
	\label{fig:size}
\end{figure} 

The visual comparison of point clouds upsampling and surface reconstruction (using~\cite{kazhdan2013screened}) is shown in Figure~\ref{fig:up_visual}, in which (a) is 2,048 random input points, (b) is the ground truth points uniformly-sampled from the original meshes of testing models, (c-f) are state-of-the-art methods PU-Net~\cite{yifan2019patch}, 3PU~\cite{yifan2019patch} and PU-GAN~\cite{li2019pu}, and (d) is our PU-EVA. The expanded features of PU-Net~\cite{yifan2019patch} are too similar to the inputs, which affects the upsampling quality, as shown in Figure~\ref{fig:up_visual} (c). Comparing with (c-e), the proposed PU-EVA produces more uniform point clouds with less noise. According to the blown-up views of upsampled results, PU-EVA renders fine-grained details better, \eg ox's ear (top) and statue's leg (bottom). The reconstructed surfaces are smoother with less wrinkles or bulges while maintain the complex structures of the original shape. 

Figure~\ref{fig:diff_r} illustrates the upsampling results of the proposed PU-EVA with various upsampling rate, in which (a) is 2,048 random input points, (b) is 8,192 ground truth points, (c-f) are the upsampling results of $R$ = 2, $R$ = 6, $R$ = 7, and $R$ = 8, respectively. Note that all the upsampling results are obtained in one-time training with $R$ = 4, and the upsampling results of rates bigger than the training rate still create point clouds fidelity to the ground truth. The proposed PU-EVA interpolates new points based on edge vectors to decouple the upsampling rate with the network architecture, making it to be a flexible-scale upsampling method.

\subsection{Ablation study}
\begin{figure}[t]
	\begin{center}
		\includegraphics[width=1.1\linewidth]{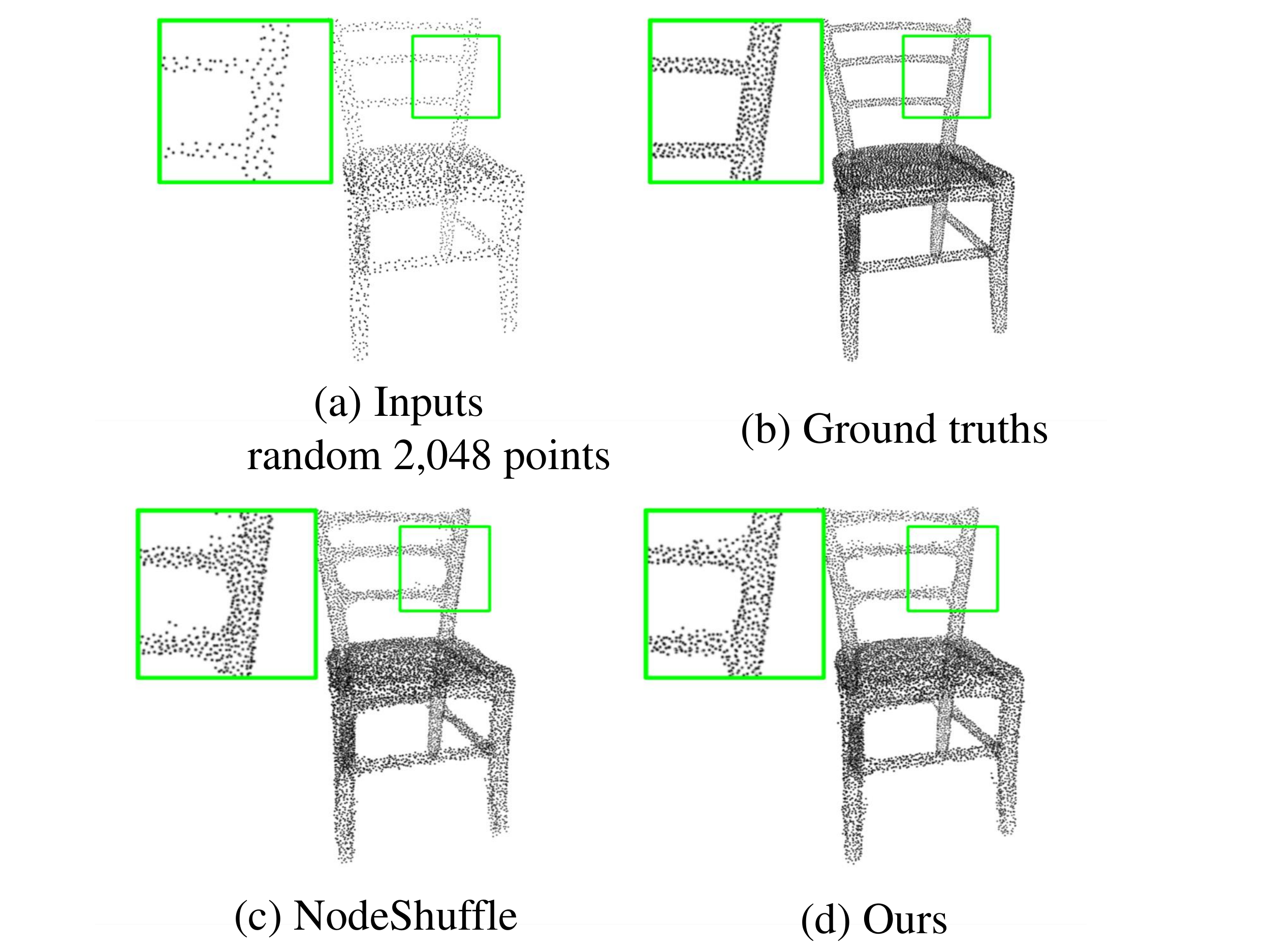}
	\end{center}
	\caption{Qualitative comparison of upsampling unit with NodeShuffle~\cite{qian2019pu}.}
	\label{fig:abl}
\end{figure}

\begin{figure*}[t]
	\begin{center}
		\includegraphics[width=1\linewidth]{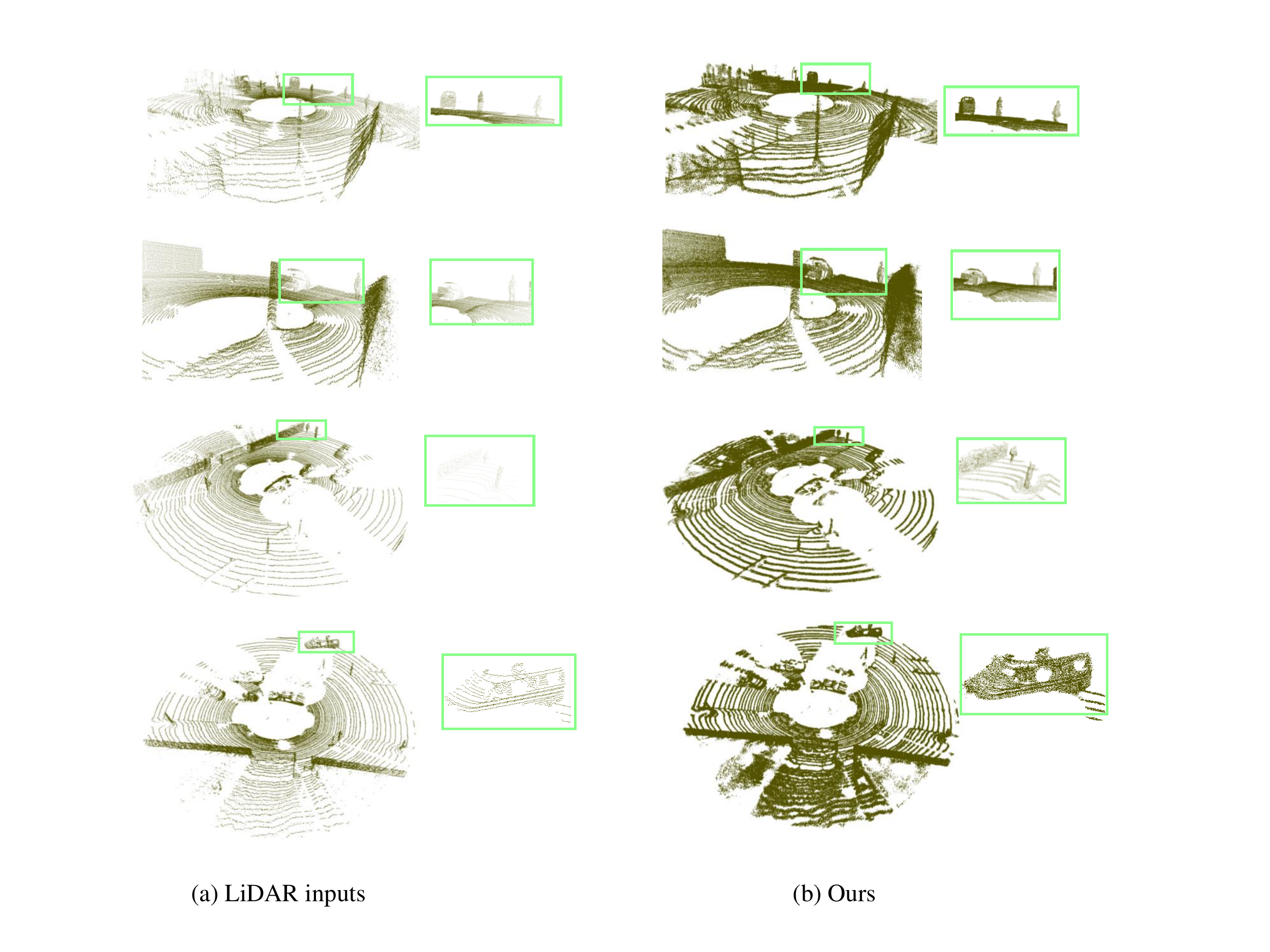}
		\caption{Real-world testing by applying PU-EVA to LiDAR-scanned street scenes from Kitti~\cite{geiger2013vision}.}
		\label{fig_kitti}
	\end{center}
	\vspace{-0.5cm}
\end{figure*}
To justify the approximation error is within the second-order term of Taylor's Expansion, the performances of PU-EVA with and without it are compared. The comparison is based on $4 \times$ upsampling rate, various input resolutions are evaluated, containing sparse points (256), medium points (2,048), and dense points (4,096), respectively. The quantitative results are shown in Table 4. The upsampling results of PU-EVA with the second-order error of Taylor's Expansion declines both CD and HD, indicating its ability to drive upsampling points approaching the underlying surface. Nevertheless, please note that even without the second-order error term, the network still accomplishes comparable upsampling performance.

To explore the efficacy of the proposed EVA upsampling unit, we conduct ablation study to compare with different upsampling units. Specifically, direct point feature duplication upsampling, and the best upsampling unit in PU-GCN~\cite{qian2019pu}, named NodeShuffle, are adopted for comparison. The direct point feature duplication follows the network architecture of the Feature Expansion module in PU-Net~\cite{yu2018pu}, while keeps rest of network the same as PU-EVA. Table 5 summarizes quantitative comparison in terms of surface deviations. According to all three metrics, the proposed EVA upsampling unit achieves the best performance with the lowest deviation from surface consistently across the input point resolutions, convincing our upsampling unit encourages points to approximate the underlying surface better. Besides, we visualize the upsampling results of EVA upsampling and NodeShuffle in~\cite{qian2019pu}. As shown in the blown-up views in Figure~\ref{fig:abl}, EVA upsampling unit achieves better upsampling results in term of fine-grained details.

\subsection{Robustness test}
\textbf{Robustness against varying levels of input noise.} 
We evaluate the robustness of proposed PU-EVA by comparing with the state-of-the-art method, PU-GAN~\cite{li2019pu}, against varying levels of input noise. Figure~\ref{fig:noise} and Table 3 demonstrate the comparison of upsampling results with additive Gaussian noises, the different additive noise levels of input point clouds are $\sigma=$ 0, 0.01, 0.02, from left to right. Compared with PU-GAN~\cite{li2019pu} upsampled point clouds, the proposed PU-EVA generates points approximating surfaces better, especially in the cases with heavier noise contamination. Furthermore, with the additive Gaussian noise levels escalating, the proposed PU-EVA shows the robustness of the upsampling result. 

\textbf{Robustness against varying sizes of input.} 
Figure~\ref{fig:size} illustrates the upsampling results with respect to varying sizes of input. We sample four sets of test samples, as shown in the upper row, which are 4,096, 2,048, 1,024, and 512 points, respectively. The bottom row shows the corresponding $4 \times$ upsampled outputs. We can witness that even for the sparsest input points in the rightmost column, the proposed PU-EVA acquires desirable performance.

\begin{figure}[t]
	\begin{center}
		\includegraphics[width=1.05\linewidth]{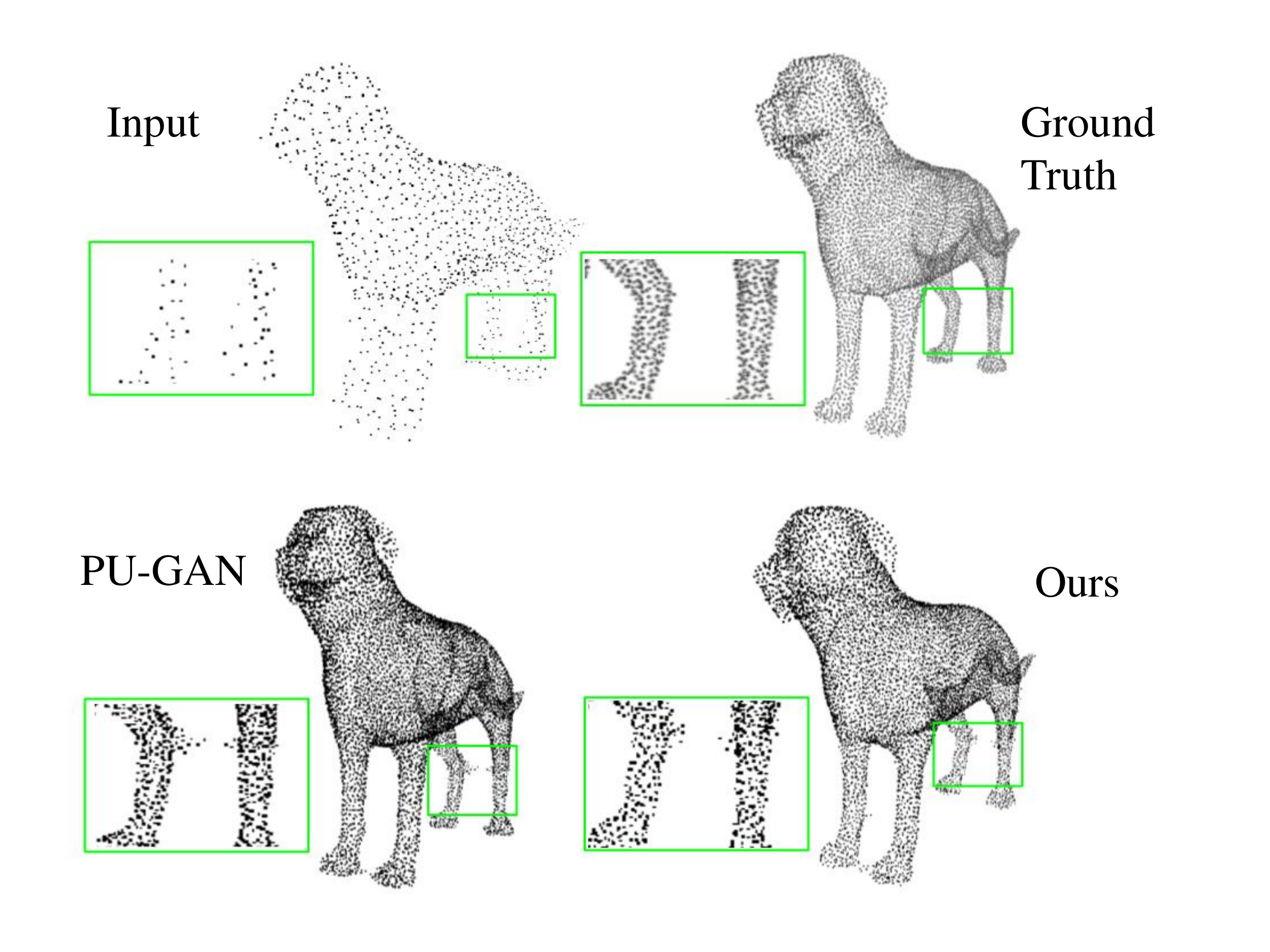}
		\caption{A typical failure case for PU-EVA.}
		\label{fig_fail}
	\end{center}
	\vspace{-0.5cm}
\end{figure}

\subsection{Real-scanned LiDAR results}
The upsampling results of PU-EVA on LiDAR-scanned street scenes from Kitti~\cite{geiger2013vision} is provided in Figure~\ref{fig_kitti}. Visualization results show that the proposed PU-EVA recovers the sparse and non-uniform input to obtain sharper object shapes of pedestrian and cyclist, etc.

\subsection{Failure case}
Although our method handles the fine-grained details well, such as the ox’s ear and the statue’s leg shown in Figure~\ref{fig:up_visual}, there are still some very challenging cases making the upsampled results fail to approximate the complex geometries accurately. The blown-up views of Figure~\ref{fig_fail} visualize details of the failure case, the possible explanation is the edge vector based affine combination method tends to over fill the local neighborhood, with the characteristics inherited from the intrinsic mechanism.

\section{Conclusion}\label{conclu}
This work is motivated to embed enriched geometric information of the underlying surface for point clouds upsampling. To this end, we first construct a similarity matrix to encode the correlations of edge vectors in local neighborhood, and propose PU-EVA to implement the significance of edge vectors for upsampling. By analyzing the linear approximation theorem, we identify the error of the edge vector based approximation are under control and verify the feasibility of the proposed network PU-EVA. Qualitative and quantitative evaluations demonstrate that the proposed PU-EVA outperforms the state-of-the-arts by generating smoother and more uniform dense point clouds maintaining more fine-grained details. In future research, we are interested in incorporating the upsampling results to bolster the performance of downstream works.

{\small
\bibliographystyle{ieee_fullname}
\bibliography{egbib}
}

\end{document}